\documentclass[preprint,12pt]{elsarticle}

\usepackage{amssymb}
\usepackage{lineno}
\usepackage{multicol}
\usepackage{multirow}
\usepackage{empheq}
\usepackage{verbatim}
\usepackage{parskip}
\usepackage{geometry}
\usepackage[pdftex]{hyperref}
\usepackage[table]{xcolor}%
\usepackage{color, colortbl}
\usepackage{slashbox}
\journal{Expert Systems with Applications}
\begin{document}

\begin{frontmatter}



\title{A Hybrid Two-layer Feature Selection Method Using Genetic Algorithm and Elastic Net}


\author{Fatemeh Amini$^a$, Guiping Hu$^{b,*}$}

\address{$^a$\small{Department of Industrial and Manufacturing Systems Engineering, Iowa State University, Ames, IA} 50011\\ email address: famini@iastate.edu \\phone number: (515) 817-3737}

\address{$^{*,b}$\textbf{Corresponding author:} \small{Department of Industrial and Manufacturing Systems Engineering, Iowa State University, Ames, IA} 50011\\ email address: gphu@iastate.edu \\phone number: (515) 294-8638}

\vspace{-0.5cm}

\begin{abstract}
Feature selection, as a critical pre-processing step for machine learning, aims at determining representative predictors from a high-dimensional feature space dataset to improve the prediction accuracy. However, the increase in feature space dimensionality, comparing to the number of observations, poses a severe challenge to many existing feature selection methods with respect to computational efficiency and prediction performance. This paper presents a new hybrid two-layer feature selection approach that combines a wrapper and an embedded method in constructing an appropriate subset of predictors. In the first layer of the proposed method, Genetic Algorithm(GA) has been adopted as a wrapper to search for the optimal subset of predictors, which aims to reduce the number of predictors and the prediction error. 
As one of the meta-heuristic approaches, GA is selected due to its computational efficiency; however, GAs do not guarantee the optimality. To address this issue, a second layer is added to the proposed method to eliminate any remaining redundant/irrelevant predictors to improve the prediction accuracy. Elastic Net(EN) has been selected as the embedded method in the second layer because of its flexibility in adjusting the penalty terms in regularization process and time efficiency. This hybrid two-layer approach has been applied on a Maize genetic dataset from NAM population, which consists of multiple subsets of datasets with different ratio of number of predictors to number of observations. The numerical results confirm the superiority of the proposed model.
\end{abstract}


\end{frontmatter}

%\linenumbers

\section{Introduction}
\label{S:1}

Advances in information technology has led to increasingly large datasets in both number of instances and number of predictors, such as applications in text mining and bioinformatics (Guyon \& Elisseeff, 2003). One significant problem for prediction with high dimensional data is that the number of predictors exceeds the number of observations (Yu \& Liu, 2003). In these situations, some of predictors maybe redundant, irrelevant, and harmful for the model training (Cilia, De Stefano, Fontanella, \& Scotto di Freca, 2019; Xue, Zhang, Browne, \& Yao, 2015. Redundant predictors provide information that is already represented with other predictors, while irrelevant predictors do not contribute to model training (Welikala et al., 2015). In fact, these predictors unnecessarily increase the computation time and deteriorate the performance of the classification/regression models (Lin, Huang, Hung, \& Lin, 2015; Oztekin, Al-Ebbini, Sevkli, \& Delen, 2018). Thus, extracting a smaller subset of predictors with most relevant predictors would be essential since it saves time in data collection and computation, and avoids overfitting problem in the prediction models (Aytug, 2015). Feature selection methods have been introduced to filter out the irrelevant and redundant predictors to achieve the smallest, most powerful subset of predictors in order to not only reduce the computation time, but also improve the prediction accuracy (Huang \& Wang, 2006; Lin et al., 2015).

Feature selection approaches can be categorized into three broad classes: the filter methods, the wrapper methods, and embedded methods. For filter methods, each individual predictor is evaluated with a statistical performance metric and then ranked according to its performance indicator. Truncation selection is then applied to select the top performing features before applying machine learning algorithms. Filter methods serve as a preprocessing step, since they do not consider the complex interactions between predictors and are independent of learning algorithms (Guyon \& Elisseeff, 2003; Hu, Bao, Xiong, \& Chiong, 2015). These methods are computationally efficient; however, they suffer from getting stuck in local optimum because the complex interactions among predictors may have been ignored (Cheng, Sun, \& Pu, 2016; Hong \& Cho, 2006; Welikala et al., 2015). The second class, wrapper methods, incorporate prediction models into a predetermined objective function that evaluates the appropriateness of the predictor subsets through an exhaustive search (Kabir, Islam, \& Murase, 2010). Although wrapper methods consider the interaction among predictors, they are not as computationally efficient as filter methods because of the larger space to search (Cilia et al., 2019; Hall, 1999; Hu et al., 2015; Kabir et al., 2010). The issue arises that evaluating all possible $2^P$ predictor combinations is neither effective nor practical in terms of computation time, especially when the number of predictors, $P$, gets larger (Cilia et al., 2019; De Stefano, Fontanella, \& Marrocco, 2008; Lee et al., 2002; Peng, Long, \& Ding, 2005). Feature selection is among NP-hard problems in which the search space grows exponentially as the number of predictors increases (Hu et al., 2015; Jeong, Shin, \& Jeong, 2015). The third class, embedded methods, are more efficient than wrappers, since they incorporate feature selection as part of the training process and select those features which contribute the most to the model training (Guyon \& Elisseeff, 2003). Regularization methods, also called penalization methods, are the most common embedded methods. These methods would push the model toward lower complexity by eliminating those predictors with coefficients less than a threshold. The basic assumption of regularization methods is the linear relation between predictors and response variable, which may not hold especially in high dimensional dataset.

To avoid the aforementioned shortcomings of the existing feature selection methods, a two-layer feature selection method has been proposed in this study. The proposed method is a hybrid wrapper-embedded approach, which complements wrapper and embedded methods with their inherent advantages. For the wrapper part, a population-based evolutionary algorithm (the GA), has been adopted in the first layer of the proposed method due to the efficiency in searching process. It can achieve good performance as well as avoid exhaustive search for the best subset of predictors. This reduces the computation time of the wrapper component, while finds near optimal solution through an efficient process. However, as a meta-heuristic algorithm, there is no guarantee finding the optimal solution, therefore, in the second-layer, an embedded method is applied on the reduced subset of predictors to eliminate those remaining irrelevant predictors (Jeong et al., 2015). The assumption of linearity between reduced subset of predictors and response variable is much relaxed than linearity assumption among the full original predictors and response variable. In the implementation of this proposed two-layer feature selection scheme, Elastic Net (EN) is selected as the training model because of its flexibility in adjusting the penalty terms in regularization process and time efficiency.

\par The rest of the paper is organized as follows. Section 2 describes the related literature,  the motivations of this study, along with the contributions of this paper. Section 3 provides  background on the mathematical model of the GA, EN method and the proposed hybrid approach. A description on the case study which the proposed method has been applied to is also covered in Section 3.
Section 4 explains the detailed experimental setting, discusses the results of the hybrid Genetic Algorithm-Elastic Net(GA-EN) method and compares the results with selected counterparts in terms of the prediction accuracy. Finally, section 5 concludes this study and suggests future research directions.

\section{Related Work}

Genetic Algorithms, as a meta-heuristic search strategy, have mainly been adopted to find the optimal hyper-parameters for machine learning algorithms. A modified genetic algorithm, known as a real-value GA, was constructed to find the optimal parameters for a Support Vector Machine (SVM) algorithm. The algorithm was then applied to predict aquaculture quality (Liu et al., 2013). Similarly, the set of optimal parameters for both SVM and Random Forest (RF) have been found using GA. The SVM and RF models were then applied to construct a fire susceptibility map for Jiangxi Province in China (H. Hong et al., 2018). 

Recently, the applications of GA are going beyond the hyper-parameter tuning of prediction models. They have been adopted as a search strategy inside the feature selection methods because of their ability to avoid exhaustive search that reduces high dimensional feature spaces. So far, many studies have combined GA with machine learning algorithms to improve the prediction accuracy especially in classification problems. (Cerrada et al., 2016) implemented GA to reduce the feature space to construct a more efficient RF model that predicts multi-class fault diagnosis in spur gears.
As (Oztekin et al., 2018) illustrated, GA was combined with three different machine learning methods, K-Nearest Neighbor(KNN), SVM, and Artifitial Neural Network (ANN) to improve the prediction accuracy of the patient quality of life after lung transplantation. Although the GA-SVM model outperforms both the GA-KNN and GA-ANN approach, these last two models still yield high prediction accuracies. 
Among the hybrid methods of different machine learning with GA, deep synergy adaptive-moving window partial least square-genetic algorithm (DSA-MWPLS-GA), was designed to obtain accurate predictions of common properties of coal (Wang et al., 2019). 
Additionally, (Cheng et al., 2016) combined a GA with a Successive Projections Algorithm to select the most relevant wavelengths. The most five important wavelengths were then used to establish Least-Squares Support Vector Machine (LS-SVM) and Multiple Linear Regression (MLR) models in order to predict drop loss in grass carp fish.
This is further evidenced by (Cornejo-Bueno, Nieto-Borge, García-Díaz, Rodríguez, \& Salcedo-Sanz, 2016) in which a new hybrid feature selection method was proposed. The method combines Grouping Genetic Algorithm with an Extreme Learning Machine approach (GGA-ELM). The GGA was used as a search strategy to find the ideal subsets, while the ELM was implemented as the GGA’s fitness function to evaluate the candidate subsets. The GGA-ELM model yielded a significantly smaller RMSE value than the ELM model using all features, validating that combining feature selection approaches can improve overall model performance. The model was then applied on marine energy datasets to predict the significant wave height and energy flux.
Moreover, most of hybrid approaches have been applied on classification problem and not much attention has been devoted to regression problems. This serves as one of the major motivations in this study.

It should be noted that, GAs can only be combined with supervised learning algorithms with a response variable. For datasets without response variable, clustering and classifying based on the feature space should be applied before implementing GAs. (Sotomayor, Hampel, \& Vázquez, 2018) firstly, applied K-means clustering approach to classify the water station into two types, based on their associated water quality. A hybrid model was then developed with K-nearest neighbor and GA to reduce the dimension of feature space and achieve higher prediction accuracy.

One of the most common concerns on high-dimensional datasets is that models are prone to overfitting, which is aggravated as the ratio of predictors to observations increases (Guyon \& Elisseeff, 2003). It can be observed that the performance of a feature selection mechanism can be improved if it is carefully designed in conjunction with another filter or wrapper approach, as it will further reduce the feature space and facilitate the design of a more efficient and accurate prediction model. Therefore, two-layer feature selection approaches have been proposed to extract the best subset from the selected predictors obtained from the first layer feature selection. (Hu et al., 2015) proposed a hybrid filter-wrapper method that uses a Partial Mutual Information (PMI) based filter method as the first layer to remove the unimportant predictors. Once the dimensions of feature space are reduced, a wrapper process consisting of a combination of a SVM and the Firefly Algorithm (FA), which is a population-based meta-heuristic technique, was then applied on the reduced feature space. 
However, since filter methods, such as the PMI approach do not take into account the possible dependencies/interactions among predictors, the performance when applied for high dimensional feature spaces is not satisfactory. This is due to the fact that two factors may be independently counted as irrelevant and/or redundant predictors, when keeping both in the model could result in performance gain.
In this paper, the proposed algorithm adopts a wrapper, as its first layer of feature selection and an embedded method, EN regularization algorithm, as the second layer in order to reduce feature space dimension while improving the prediction accuracy.

\par The contributions of this study can be summarized as follows. Firstly, unlike most existing studies, which focused on the combination of a GA with a machine learning algorithm for classification problems, our proposed hybrid model approach focuses on regression problems. 
Secondly, an embedded method, EN, is combined with a wrapper method, GA, which have not been addressed in previous studies.
Thirdly, by defining a more complex fitness function for the GA, the optimal subset will be achieved for the the smallest number of predictors with the lowest root mean square error of prediction ($RMSE$).

%\begin{figure}[h]
%\centering\includegraphics[width=100mm , scale =1.5]{svm-ga.PNG}
%\caption{GA-EN process}
%\end{figure}\footnote{Genetic Algorithm-Elastic Net}
\section{Methods and Materials }

This section describes the proposed two-layer feature selection method. In the first layer, a wrapper has been designed to select the best subset of predictors with lowest prediction error while includes as few predictors as possible. This is done with a genetic algorithm based search strategy. In the second layer, EN has been applied to further eliminate the remaining redundant/irrelevant predictors to improve the prediction accuracy, using the best subset of predictors outputted from the first layer. Additionally, the case study adopted to validate the proposed method has been described in this section.

\subsection{Elastic Net Regularization Method}
EN regularization is a modification of the multiple linear regression approaches designed to solve high-dimensional feature selection problems (Fukushima, Sugimoto, Hiwa, \& Hiroyasu, 2019). Using two penalty terms (L1-norm and L2-norm), the EN selects variables automatically and performs continuous shrinkage to improve the prediction accuracy. This method works like a stretchable fishing net which keeps ``all big fish'', i.e., important predictors, and eliminate those irrelevant ones (Park \& Mazer, 2018; Zou \& Hastie, 2005).

\par Suppose that we have \textit{p} predictors denoted by $x_{1}, \ldots, x_{p}$, an estimate of the response variable $y$ can be written as $\hat{y}=\beta_{0}+\beta_{1}x_{1}+\ldots+\beta_{p}x_{p}$, based on linear regression. The coefficients ($\beta$ = [$\beta_{0},\ldots,\beta_{p}$]) are calculated by minimizing the sum of the squares of the error residuals (Eq.(\ref{SSE})). In the case where the dimensions of the feature space are greater than the number of observations, the coefficients are calculated by minimizing the L function (Eq.(\ref{L function})) instead of minimizing SSE (Wei, Chen, Song, \& Chen, 2019):

\begin{equation}
\label{SSE}
SSE = ||Y-X\beta||^2
\end{equation}

\begin{equation}
\label{L function}
L =  SSE + \alpha \rho ||\beta||_{1} + \alpha (1-\rho) ||\beta||^2
\end{equation}

\vspace{10 pt}

Where $||\beta||_{1}$ and $||\beta||^2$ are calculated with Eqs.(3 and 4).

\begin{equation}
||\beta||_{1} = \sum_{p=1}^{P} |\beta_{p}|
\end{equation}

\begin{equation}
||\beta||^2 = \sum_{p=1}^{P} \beta_{p}^2
\end{equation}

The degree to which model complexity is penalized is controlled by weighting terms $\alpha$ and $\rho$. As the outcome of the Elastic Net is affected by $\alpha$ and $\rho$, tuning them should be done within the learning process. Two special cases for EN are when $\rho = 1$ and $\rho = 0$. When $\rho = 1$, EN regression is reduced to \textit{LASSO}, which aims to reduce the number of non-zero linear coefficients to zero in order to create a sparse model. When $\rho = 0$, EN regression is reduced to ridge regression, which allows the model to include a group of correlated predictors to remove the limitation of number of selected predictors (Chen, Xu, Zou, Jin, \& Xu, 2019; Park \& Mazer, 2018; Wei et al., 2019). It is shown that as EN is able to select a subset of highly correlated features, it avoids the shortcoming of high-dimensional feature selection when solely using \textit{LASSO} or ridge regression methods (Zou \& Hastie, 2005).

\subsection{Genetic Algorithms}
GAs are one of the meta-heuristic search methods that implement a probabilistic, global search process that emulates the biological evolution of a population, inspired by Darwin’s theory of evolution (Cheng et al., 2016; Welikala et al., 2015). GAs are powerful tools for achieving the global optimal solution of large-scaled problems (Cerrada et al., 2016; Liu et al., 2013). The GA process can be described in these steps:

\begin{enumerate}
\item  Individual encoding: Each individual is encoded as binary vector of size $P$, where the entry $b_i=1$ states for the predictor $p_i$ that is defined for that individual, $b_i=0$ if the predictor $p_i$ is not included in that particular individual ($i=1,\dots, P$)(Cerrada et al., 2016).
\item Initial population: Given the binary representation of the individuals, the population is a binary matrix where its rows are the randomly selected individuals, and the columns are the available predictors. An initial population with a predefined number of individuals are generated with random selection of $0$ and $1$ for each entry (Cerrada et al., 2016).

\item Fitness function: the fitness value of each individual in the population is calculated according to a predefined fitness function (Welikala et al., 2015). The highest fitness value would be assigned to the individual with lowest prediction error while includes fewest predictors.
\item Applying genetic operators to create the next generation.
\begin{itemize}
    \item Selection: The elite individuals, those with highest fitness values, are selected as parents to produce children through crossover and mutation processes. In this study, instead of selecting all parent from the highest qualified individuals, a random individual will also added to the parent pool in order to maintain generational diversity. Each pair of parents produces a number of children to create the next generation, which has the same size as the initial population. To stabilize the size of each generation, Eq. (\ref{size}) should be satisfied.\\
    \begin{equation}
    \frac{\# of BS + \# of RS}{2} * \# of children = \mbox {initial population size}. 
    \label{size}
    \end{equation}
    
    \vspace{5pt}
    
    BS is the best selected individuals, while RS is the randomly selected individuals.

    \item Crossover: It is a mechanism in which new generation is created by exchanging entries between two selected parents from the previous step. A single point crossover technique has been used in this paper (Liu et al., 2013; Welikala et al., 2015).
    \item Mutation: This operation is applied after crossover and determines if an individual should be mutated in next generation or not and makes sure no predictors has been removed from GA's population permanently (Brown \& Sumichrast, 2005; Liu et al., 2013).

\end{itemize}
\item Stop criteria: Two stop criteria are widely used in GAs. The first one, used in this study, is reaching the maximum number of generations. The other one is the lack of fitness function improvement in two successive generations (Cheng et al., 2016). Steps 2 and 3 are performed iteratively until the stop criterion is met.

\end{enumerate}

\subsection{Proposed GA-EN feature selection approach}
The proposed feature selection method has two layers. In the first layer, GA has been implemented to reduce the search space to find the best subset of predictors, thus a small subset of predictors can be identified to reduce the computational cost and improve the prediction accuracy. In the second layer, EN regularization method is adopted to eliminate those remaining redundant predictors in the feature space given in the first layer. The reason for choosing EN as the regressor is that not only the EN makes use of shrinkage to reduce the high-dimensional feature space, but also it tends to outperform other models in regression problems. Thus, the probability of having redundant/irrelevant predictors in the final model would decrease, resulting in a prediction model without any significant sign of overfitting. 

\begin{figure}[h!]
\centering
\includegraphics[width=150mm , scale =2]{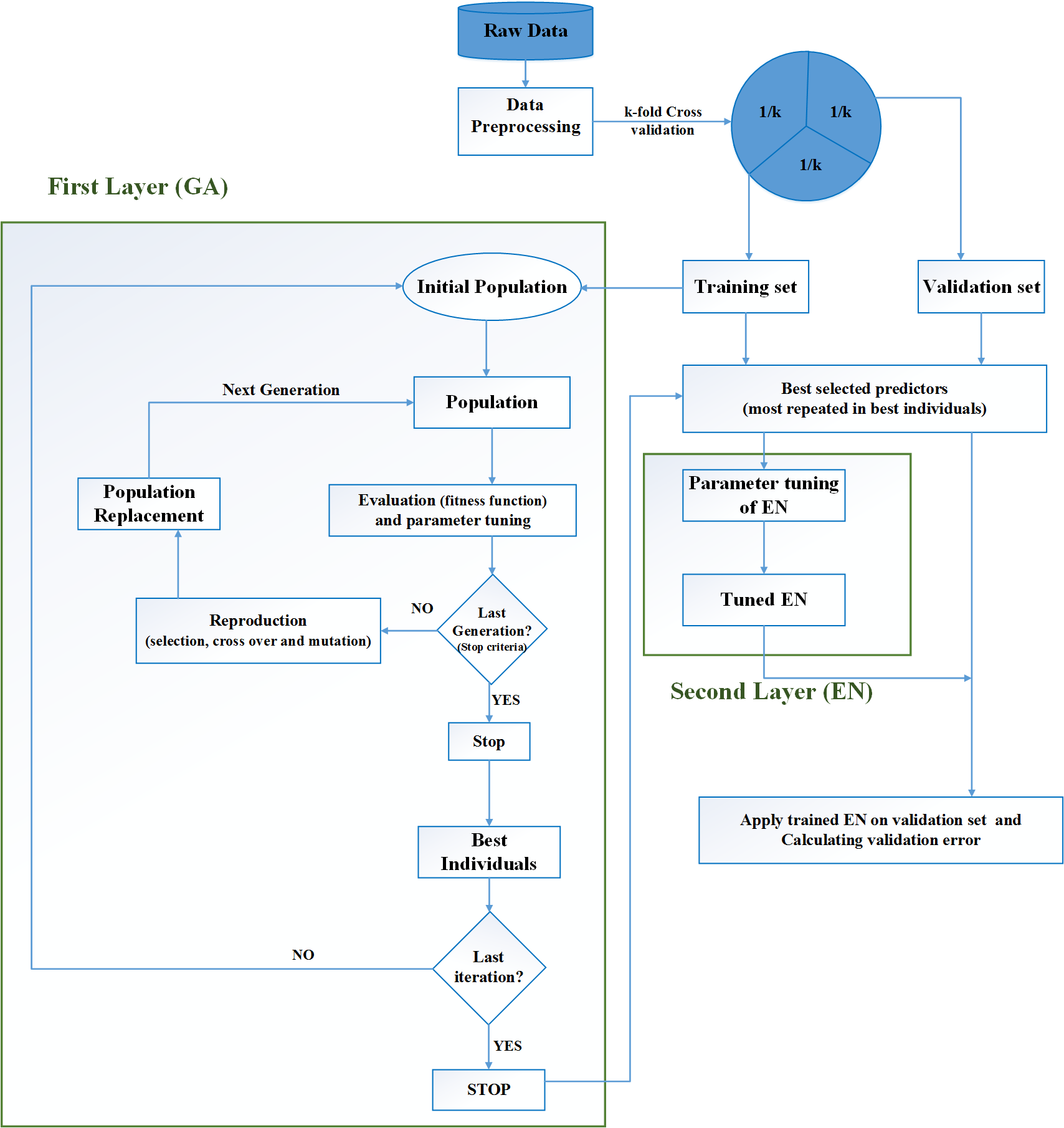}
\caption{Flowchart for the proposed GA-EN approach}
\label{fig:GA-EN}
\end{figure}

The architecture of the proposed two-layer feature selection method is described in Figure \ref{fig:GA-EN}. Following pre-processing the data, using k-fold cross validation technique, data is splitted into \textit{k} folds in which $k-1$ folds is considered as the training set and 1 fold as the the validation set. This procedure is repeated \textit{k} times such that each fold will be used once for validation. Averaging the $RMSE$ over the \textit{k} trials would provide an estimation of the expected prediction error (Eq. (\ref{rmse})), which is the performance evaluation metric. The main idea behind the k-fold cross-validation is to minimize any potentail bias of random sampling of training and validation data subset (Oztekin et al., 2018).

In the first layer of the proposed method, training set is fed into GA to search for the best subset of predictors. Throughout the GA search procedure, after building the initial population, individuals are ranked according to their fitness values and the highest ranked ones are more likely to be selected in the selection process to create the next generation. The GA runs multiple times, and each iteration outputted a best subset of predictors. Then, the predictors that have been repeated frequently in the best subset of predictors in each iteration of GA, would be included in the final subset of predictors. Therefore, the most important predictors can be identified as those repeated more often in the best subset given by GA. A threshold is considered to specify how often a predictor should be repeated in the best subset of GA to be included in the final subset of predictors given in the first layer of the proposed method. The higher this threshold is defined, the stricter the model in selecting important predictors.

\begin{equation}
 RMSE_{CV} =  \frac{1}{\#folds}\sum_{1}^{\#folds}\frac{1}{n}\sum_{i=1}^{n}\sqrt{(y_{i} -\hat{y_{i}})^{2}}
 \label{rmse}
\end{equation}

\vspace{0.5cm}

In the second layer, the EN was applied to the new dataset composed of the predictors selected by the GA to eliminate those redundant predictors which are not eliminated by GA. Elastic Net is a powerful tool that helps further reduce the number of predictors selected in the first layer, and, thus, improving the performance of the model; however as its performance significantly depends on the hyper-parameters, $\alpha$ and $\rho$ , they are required to be tuned through the training process. Finally, the tuned model is ready to be evaluated on the validation set. It should be noted that the hyper-parameters tuning are calculated through a k-fold cross-validation process, as well.

The applied fitness function of GA which incorporates EN as the regression model to evaluate the fitness value associated with the individuals is shown as Eqs. (6-11).

\begin{eqnarray}
    \label{fitness function}
  \min_{\hat{y_{i}}, n_{p}} & w_{r}* r_{RMSE}  + w_{p}*n_{p} & \\
  \label{fitness function1}
  s.t. & 0 \leq r_{RMSE} =\frac{1}{\overline{y}}(\frac{\sum_{i=1}^{n}\sqrt{(y_{i}-\hat{y_{i}})^2}}{n}) \leq 1 &\\
  \label{fitness function2}
  & 0 \leq n_{p} =\frac{\sum_{p=1}^{P} f_{p}}{P} \leq 1 & \\
  \label{fitness function3}
  &  w_{r}+ w_{p} = 1 &\\
  \label{fitness function4}
  & w_{r}, w_{p} \geq 0 \\
  & f_{p} \in \{0,1\}
  \label{fitness function5}
\end{eqnarray}

 The objective function in Eq.(\ref{fitness function}) minimizes the prediction error, $RMSE$ and the number of predictors used, as the goal is to achieve higher prediction accuracy with minimum number of predictors included. Eq.(\ref{fitness function1}) defines the relative $RMSE$, where $y_{i}$ and $\hat{y_{i}}$ are the actual and prediction values of the response variable, respectively. The number of selected predictors is defined as $n_{p}$ in Eq.(\ref{fitness function2}), where $f_p$ is a binary variable that denotes if predictor $p$ is included in a particular individual or not. $w_{r}$ and $w_{p}$ in Eqs.(\ref{fitness function3} and \ref{fitness function4}) are the importance of the prediction error and number of selected predictors,respectively, which sum up to 1. The best values of $w_{r}$ and $w_{p}$ are determined regarding the purpose of the project, by considering the minimum cross-validation error achieved. If reducing the number of predictors is preferred to be more important than reducing $RMSE$, then $w_{r} < w_{p}$ and vice versa.

\vspace{5pt}

\subsection{Data Description and Pre-processing}

\begin{figure}[h!]
\centering
\includegraphics[width=100mm , scale =1]{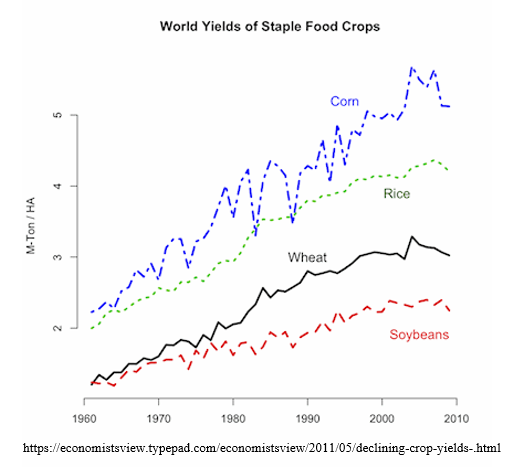}
\caption{Crops production in US}
\label{fig:corn}
\end{figure}

Motivated by the importance of agricultural system in food production, particularly Maize plants in US (Figure \ref{fig:corn}), a case study on Maize traits prediction has been carried out to demonstrate the outperforming of the the proposed hybrid feature selection method. In this case study, the SNPs (Single Nucleotide Polymorphisms) data of Maize parents are collected to predict their expression level (RNA-seq) information. The US-NAM parents data is used in this paper which is publically available  at NCBI SRA under SRA050451 (shoot apex) and SRA050790 (ear, tassel, shoot, and root) and at NCBI dbSNP handle PSLAB, batch number 1062224.

This dataset contains the expression level information for about 6000 genes of 27 Maize parents in addition to about 4 million SNP data associated with those parents. As a biological pre-processing step to reduce the number of SNPs, the co-Expression quantitative trait loci (eQTL) analysis has been conducted to identify the most important SNPs related to each gene. SNPs importance level is determined by a predefined distance around each gene and those included in this distance are counted as important SNPs (Kusmec, Srinivasan, Nettleton, \& Schnable, 2017). The shorter this distance is defined, the fewer number of SNPs would be included. The distance considered for the eQTL analysis in this study resulted in, on average, 123 SNPs for each gene.
This process reduces the number of SNPs from $\sim$4 million to $\sim$728000. Moreover, SNPs data are converted to binary representation. For missing data, a linear regression based imputation method has been implemented based on the two nearest SNPs. Thus, a prediction model is defined for each gene which aims to predict the expression level of Maize parents based in their SNPs information.

 All gene datasets contain same number of observations (27 parents) while the number of predictors (SNPs) are different. In order to validate the prediction accuracy improvement of the proposed model on datasets with different ratio of number of predictors to number of observations, 10 gene datasets have been selected in a way that a diverse range of this ratio has been covered.

\section{Numerical Results and Analysis}
The objective of this section is to evaluate the proposed two-layer feature selection method in terms of reduction in feature space dimension and prediction accuracy. Moreover, tuning the hyper-parameters in both GA and EN should be carried out prior to model evaluation since it is expected to improve the performance of the model.

\subsection{Performance Metrics}
In this paper, due to the continuity of the response variable, \textit{relative} $RMSE_{CV}$ is considered as the performance evaluation. It is calculated through a 3-fold cross-validation process. 3-fold is chosen since 27 is dividable by 3, thus each observation will be included in just one fold at a time. \textit{relative} $RMSE_{CV}$ is calculated by Eq.(\ref{rel RMSE def}).

\begin{equation}
    \mbox{relative }RMSE_{CV} = \frac{RMSE_{CV}}{\Bar{y}}
    \label{rel RMSE def}
\end{equation}

\subsection{Hyper-parameter Tuning}

There is no universal fixed parameters for GA and as they greatly affect the GA efficiency, they need to be generally tuned to specific problems, therefore, GA parameters should be set in such a way that highest exploitation is achieved. To do that, GA is required to find a very good solution in early stages of its process. In order to increase the chance of fast improvement in the GA's response, the highest possible elitism, a limited initial population size and quite high probability of mutation have been applied (Leardi, 2000). In addition, some random individuals have been selected in each generation to keep the next generation diverse at the same time. Additionally, in order to follow the time constraint, the number of generation has to remain low (Welikala et al., 2015). Table \ref{GA-parameters} summarizes the tuned GA parameters applied in this study.

\begin{table}[h!]
\centering
\caption{Tuned GA Parameters}
\begin{tabular}{c c}
\hline
\textbf{GA Parameters} & \textbf{Values/Method}\\
\hline \hline
Initial population size & 50\\

\#of generations & 10\\

Population type & Bit string\\

\#of BS & 19\\

\#of RS & 1\\

\#of offspring & 5\\

Crossover function & single-point\\

Mutation rate & 0.05\\

\hline \hline
\end{tabular}
\label{GA-parameters}
\end{table}

\begin{comment}
It should be noted that the GA parameters in Table \ref{GA-parameters} are the average value of tuned parameters to each gene dataset, while the exact tuned parameters of GA associated with each gene dataset have been adopted for further analyses.

\end{comment}

Moreover, the weights $w_{r}$ and $w_{p}$ inside the GA’s fitness function should be tuned to each gene dataset, separately. A grid search approach has been designed to select the best weights with lowest prediction error. Four different values are considered for these weights in the grid search subset to cover all possible scenarios.

 \begin{table}[h!]
\centering
\caption {Weights in fitness function}
\begin{tabular}{c |c c c c}
\hline \hline

\textbf{Scenario} & 1 & 2 & 3 & 4\\
\hline

\textbf{$w_{r}$} & 0.15 & 0.5 & 0.85 & 1\\

\textbf{$w_{p}$}& 0.85 & 0.5& 0.15 & 0\\

\hline \hline
\end{tabular}
\label{weights}
\end{table}

\begin{figure}[hbt!]
\centering\includegraphics[width=150mm , scale =1]{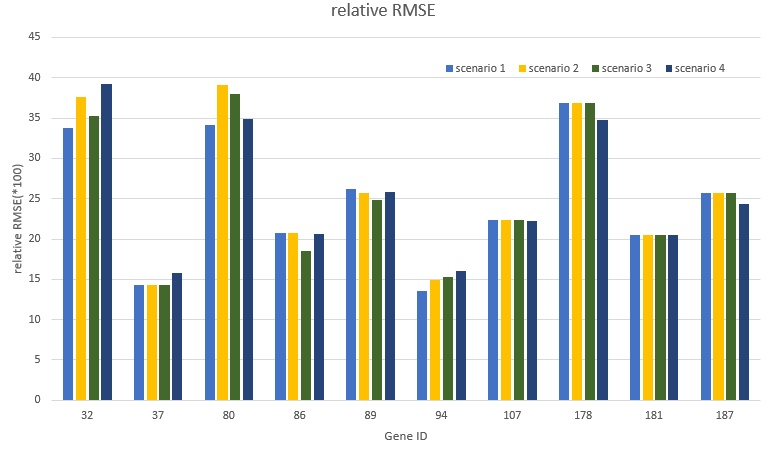}
\caption{Relative RMSE for different scenarios}
\label{fig:rel RMSE weights}
\end{figure}

Table \ref{weights} shows the different scenarios in which the higher the weight, the more emphasis is imposed on the minimization of associated term. From scenario 1 to scenario 4, more emphasis has been imposed on reducing the prediction error than decreasing the number of selected predictors. The special scenario with $w_{r} = 0, w_{p}=1 $ is not considered since the main purpose in this study is to improve the prediction accuracy and solely focusing on minimizing the number of selected predictors would not achieve this goal.
The best pair of weights with lowest \textit{relative} $RMSE_{CV}$ is then selected for each gene dataset and further analyses are implemented with the selected weights. Figure \ref{fig:rel RMSE weights} demonstrates the comparison of the \textit{relative} $RMSE_{cv}$ among all different scenarios for each gene dataset.

\begin{figure}[hb!]
\centering\includegraphics[width=150mm , scale =1]{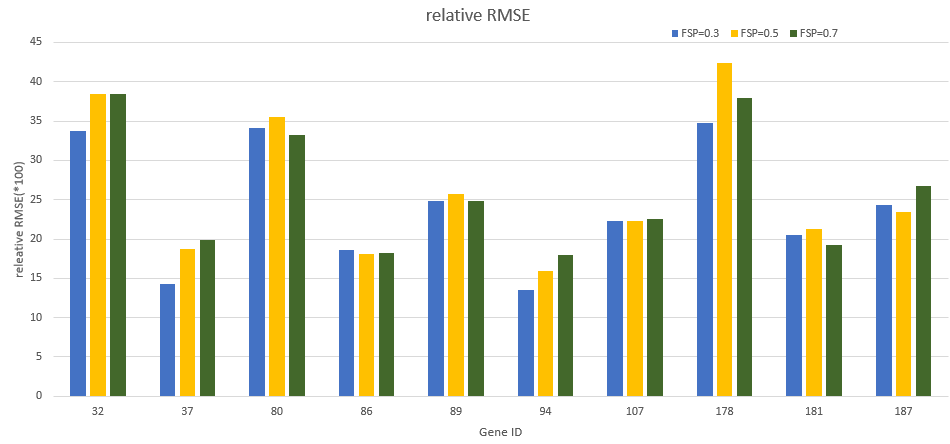}
\caption{Relative RMSE for different FSPs}
\label{fig:rel RMSE FSP}
\end{figure}

Following $w_{r}$ and $w_{p} $, the next significant parameter to be tuned is the Fraction of Selected Predictors($FSP$). It is a threshold that defines how often a particular predictor should be included in the best individuals of GA in each iteration, in order to be included in the final subset of predictors in the first layer of the proposed method. The larger this threshold is, the stricter the model is in selecting predictors.
In this study, GA is repeated 5 times and each iteration, provided us with the best individual (best subset of predictors) throughout 10 generations. To tune this parameter, three values ($0.3, 0.5$, and $0.7$) have been considered in a similar grid search approach to select the best $FSP$ in terms of lowest \textit{relative} $RMSE_{CV}$ for each gene dataset.
This grid search subset is designed in a way to incorporate low, medium, and high strictness of the method. \textit{Relative} $RMSE_{CV}$ results associated with different $FSP$ in the grid search subset for each gene dataset is illustrated in Figure \ref{fig:rel RMSE FSP}. The best $FSP$ which gives the minimum  \textit{relative} $RMSE_{CV}$ is selected for further analyses.

With $FSP$, $w_{r}$, and $w_{p}$ fixed, the EN hyper-parameters ($\rho$ and $\alpha$) should be tuned within the second layer of the proposed two-layer feature selection method. EN selects best $\alpha$ from 10 non-zero values considered in \textit{sklearn} library provided in \textit{Python}. These values are set automatically in a way that always includes values less than and greater than 1. For this case study $\alpha$ values are considered in range $(0.004,50)$. Moreover, $\rho$ would be selected from the grid search subset of $\{0.1, 0.3, 0.5, 0.7, 0.9 \}$. 
These hyper-parameters are tuned via a 3-fold cross-validation process and the average of the best values regarding each \textit{k} partitioning, along with the tuned hyper-parameters of the first layer are listed in Table \ref{gene description}.

\definecolor{Gray}{gray}{0.833}

\begin{table}[ht]
\centering
\caption{Tuned hyper-parameters of the proposed method}
\begin{tabular}{c||c c c c c c}
\hline
\textbf{Gene ID} & \textbf{Gene Shape Ratio} &\textbf{$\alpha_{best}$} &\textbf{$\rho_{best}$} & \textbf{$w_{r}$} & \textbf{$w_{p}$} & \textbf{FSP}  \\
\hline \hline
32 & 1.33 & 0.017 & 0.36 & 0.15 &0.85& 0.3\\

\rowcolor{Gray}
37 & 9.074 & 0.457 & 0.36 & 0.85 &0.15 &0.3  \\
80 & 2.89 & 30.05 &0.36 & 0.15 &0.85 & 0.7 \\
\rowcolor{Gray}
86 & 3.18 & 0.435 &0.3 &0.85 &0.15 &0.5\\

89 & 2.44 & 3.38 &0.23 &0.85 &0.15&0.3\\
\rowcolor{Gray}
94 & 7.33&0.21&0.36&0.15 &0.85&0.3\\

107 & 3.66&9.74 &0.36 & 1& 0& 0.3 \\
\rowcolor{Gray}
178 &2.63& 3.31 & 0.43 & 1 &0&  0.3 \\

181 & 10.62&3.08 &0.1& 1&0 &0.7\\
\rowcolor{Gray}
187 & 2.33&0.56&0.63&1&0&0.5\\

\hline \hline
\end{tabular}
\label{gene description}
\end{table}

Table \ref{gene description} also includes \textit{Gene Shape Ratio} which defines the ratio of number of predictors to the number of observations for each gene dataset. As most of the gene dataset contains on average about 123 predictors (SNPs), their shape ratio belongs to $(2,3)$ interval. However, it can be seen in Table \ref{gene description} that datasets whose ratio is out of this range also have been considered in this paper to validate the performance of the proposed method for datasets with different shape ratio.

\subsection{Model Validation}
The results of our numerical experiments from comparing the proposed two-layer feature selection method with following benchmarks are included in this section.

\begin{enumerate}
    \item EN (embedded method) 
    \item GA combined with linear regression (wrapper method)
\end{enumerate}
Both benchmarks are considered as single-layer feature selection methods, the first one is an embedded method, while the second one (GA-Lr) is a wrapper. Outperforming these benchmarks, it confirms that the superiority of the model is not only because of GA or EN separately, but it successfully demonstrates higher prediction accuracy because of the combination of GA and EN which designs the two-layer feature selection approach. The proposed model with tuned hyper-parameters have been evaluated through 3-fold cross validation and the performance are compared in terms of \textit{relative} $RMSE_{CV}$.

\begin{figure}[hbt!]
\centering\includegraphics[width=150mm , scale =3]{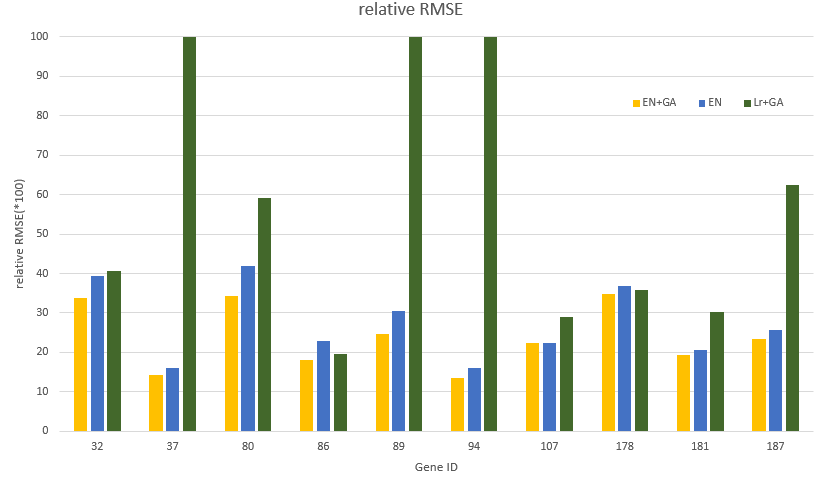}
\caption {Relative RMSE of different methods}
\label{fig:rmse_all}
\end{figure}

Figure \ref{fig:rmse_all} compares the \textit{relative} $RMSE_{CV}$ of the benchmarks with the proposed method. The results confirm that combining GA with EN that has regularization characteristics inside not only outperforms the combination of the GA with non-regularized prediction method (wrapper method), but also it does achieve better performance than applying that regularized prediction method without GA assistance(embedded method) in predicting the expression level of Maize parents. The reason behind of the outperforming of GA-EN hybrid method is that not only, the most parsimonious set of predictors along with highest level of prediction accuracy are selected in GA process in the first layer, but also the EN eliminates those insignificant and redundant predictors that still exist in the selected predictors subset in the second layer, to improve the prediction accuracy.  Moreover, it can be seen in Figure \ref{fig:rmse_all}, for some gene datasets such as gene 37, 89 and gene 94, the \textit{relative} $RMSE_{CV}$ of GA-Lr method is greater than one which means that the prediction error associated with the wrapper method is greater than the average of response variable. In these cases, embedded method in the second layer of the proposed method would be able to ignore redundant/or irrelevant predictors to improve the prediction accuracy.

Table \ref{Number of selected predictors} demonstrates the number of predictors in the original gene datasets and the number of predictors each model selects with through cross-validation. Also the \textit{relative} $RMSE_{CV}$ associated with each model with their own selected predictors are presented in Table \ref{Number of selected predictors}.
 The highlighted values show the minimum number of selected predictors and the minimum \textit{relative} $RMSE_{CV}$ for all ten gene datasets. For most gene datasets, GA-EN method demonstrates the most reduction in number of predictors along with minimum \textit{relative} $RMSE_{CV}$. In other word, this method not only reduces the dimension of the data, its complexity and required storage but also, it results in smaller prediction error. However, for some datasets such as gene 32 and 86, GA-Lr selects the smallest subset but it achieves higher prediction error.

It can importantly be said that the proposed feature selection method improves the performance of prediction model by ignoring the irrelevant and useless predictors. An important task in such a process is to capture necessary information in selecting critical predictors; otherwise the performance of the prediction model might be degraded as can be seen for gene 32 and 86. Although GA-EN selected a bulkier predictor subset compared to others, it provides lower prediction error for the these datasets. In fact, the results presented for other methods presented in Table \ref{Number of selected predictors} indicate that smallest or largest predictor subset does not guarantee the best or worst prediction accuracy.

\begin{table}[hbt!]
\centering
\caption{Result of experiment}
\scalebox{0.75}{%
\begin{tabular}{c|c|c|c|c}
\hline
%\backslashbox{\textbf{Gene ID}}{\textbf{Method}} 
\textbf{Gene ID} &\textbf{Method}  & \textbf{\# of original predictors} &{\textbf{\# of final predictors}}  & {\textbf{relative $RMSE_{CV}$}(\%)} \\

\hline \hline
\multirow{3}{*}{32} & EN+GA & &5 &\cellcolor{blue!25}{33.79}\\
&EN &36 &25.66&39.44\\
&Lr+GA &  & \cellcolor{blue!25}{2}& 40.52\\

\hline \hline
\multirow{3}{*}{37} &EN+GA & &\cellcolor{blue!25}{73.33} &\cellcolor{blue!25}{14.25}\\
& EN &245&119&15.99\\
&Lr+GA &  & 134& $>100$\\

\hline \hline
\multirow{3}{*}{80} &EN+GA & &\cellcolor{blue!25}{1} &\cellcolor{blue!25}{34.19}\\
& EN & 78&40.67&41.95\\
&Lr+GA &  & 24& 59.1\\

\hline \hline
\multirow{3}{*}{86} &EN+GA & &6 &\cellcolor{blue!25}{18.05}\\
& EN &86&22.67&22.8\\
&Lr+GA &  & \cellcolor{blue!25}{3}& 19.61\\

\hline \hline
\multirow{3}{*}{89} &EN+GA & &\cellcolor{blue!25}{8.33} &\cellcolor{blue!25}{24.78}\\
& EN &66&19.33&30.57\\
&Lr+GA &  & 19& $>100$\\

\hline \hline
\multirow{3}{*}{94} &EN+GA & &\cellcolor{blue!25}{67.33} &\cellcolor{blue!25}{13.56}\\
& EN &198&76&16.1\\
&Lr+GA &  & 95& $>100$\\

\hline \hline
\multirow{3}{*}{107} &EN+GA & &\cellcolor{blue!25}{6}&\cellcolor{blue!25}{22.25}\\
& EN &99&70&22.34\\
&Lr+GA & & 74& 29.05\\

\hline \hline
\multirow{3}{*}{178}&EN+GA & &\cellcolor{blue!25}{28.6} &\cellcolor{blue!25}{34.75}\\
& EN &71&43.67&36.82\\
&Lr+GA &  & 63& 35.7\\

\hline \hline
\multirow{3}{*}{181} &EN+GA & &\cellcolor{blue!25}{36}&\cellcolor{blue!25}{19.3}\\
& EN &287&161.67&20.56\\
&Lr+GA &  & 56& 30.12\\

\hline \hline
\multirow{3}{*}{187} &EN+GA & &\cellcolor{blue!25}{17.3} &\cellcolor{blue!25}{23.48}\\
&  EN &63 &38.67&25.67\\
&Lr+GA &  &30& 62.4\\

\hline \hline
\end{tabular}
}
\label{Number of selected predictors}
\end{table}

The comparison of results shows the effectiveness of the hybrid wrapper-embedded method in improving the prediction accuracy for regression problems.
Through the above study, we can conclude that the combination of EN with GA, including a modified fitness function in which the smallest subset of predictors with the lowest \textit{relative} $RMSE$ has been found, demonstrates higher prediction accuracy in comparison with EN and GA-EN methods in predicting the expression level of Maize plants. This hypothesis has been implemented on datasets with different ratio of number of predictors to number of observations and the results validate the superiority of the proposed model for all datasets.

\newpage

\section{Conclusions}

This paper proposed a novel two-layer feature selection approach to select the best subset of salient predictors in order to improve the prediction accuracy of regression problems. It is a two-layer method which is a hybrid wrapper-embedded method composed of GA, as the wrapper, and EN as the embedded method. In the first layer of GA-EN method, GA searches for the smallest subset of predictors with minimum prediction error. It can reduce the computation time of finding the best subset of predictors by avoiding the exhaustive search through all possible subsets. In the second layer, adopting the best subset of predictors outputted from GA, EN has been applied to eliminate the remaining redundant and irrelevant predictors. The regularization approach within the EN removes predictors with no significant relationship with the response variable. Therefore, the main contribution of this paper lies in combining a regularized learning method with GA to achieve higher prediction accuracy dealing with regression problems.

 The proposed hybrid feature selection model has been applied on real dataset of Maize genetic data which has multiple subsets of high dimesional feature space datasets with different number of predictors. Based on the numerical results, the hybrid wrapper-embedded (GA-EN) method that consists of two layer of feature elimination process, results in smaller root mean square error for all datasets with different feature space dimension, compared to the embedded (EN) method and the wrapper (GA-Lr).
 The outcome of the present study revealed that combining a wrapper and an embedded feature selection method particularly, GA and EN, would reduce the dimension of feature space by more than eighty percent on average,  without negatively affecting accuracy.

This study is subject to few limitations which suggest future research directions. Firstly, this model selects the best $w_{r}$, $w_{p}$, and $FSP$ from discrete subsets due to insufficient computational capacity and time limitation. It can be addressed in future research by letting the model select the best value of them from the continuous interval of $(0,1)$ which may improve the prediction accuracy.
Secondly, although GA is more effective than exhaustive search, large number of evaluation existed in GA, leads to high computational cost. To address this issue in the future studies, an effective representation that can reduce the the dimensionality of search space can be adopted.
Thirdly, the proposed method can be applied on datasets with different nature from what have been analyzed in this study, in terms of feature space dimension, type of the response variable, and etc. These should be reserved as future research topics.

\newpage
%\printbibliography

%\bibliography{Mahsa_Hu_final} 
%\bibliographystyle{ieeetr}



%% References




\section*{References}
\vspace{-1cm}

\begin{itemize}
\item Aytug, H. (2015). Feature selection for support vector machines using Generalized Benders Decomposition. European Journal of Operational Research, 244(1), 210–218.
\item Brown, E. C., \& Sumichrast, R. T. (2005). Evaluating performance advantages of grouping genetic algorithms. Engineering Applications of Artificial Intelligence, 18(1), 1–12.
\item Cerrada, M., Zurita, G., Cabrera, D., Sánchez, R.-V., Artés, M., \& Li, C. (2016). Fault diagnosis in spur gears based on genetic algorithm and random forest. Mechanical Systems and Signal Processing, 70–71, 87–103.
\item Chen, W., Xu, C., Zou, B., Jin, H., \& Xu, J. (2019). Kernelized Elastic Net Regularization based on Markov selective sampling. Knowledge-Based Systems.
\item Cheng, J.-H., Sun, D.-W., \& Pu, H. (2016). Combining the genetic algorithm and successive projection algorithm for the selection of feature wavelengths to evaluate exudative characteristics in frozen–thawed fish muscle. Food Chemistry, 197, 855–863. 
\item Cilia, N. D., De Stefano, C., Fontanella, F., \& Scotto di Freca, A. (2019). Variable-Length Representation for EC-Based Feature Selection in High-Dimensional Data BT  - Applications of Evolutionary Computation (P. Kaufmann \& P. A. Castillo, eds.). Cham: Springer International Publishing.

\item Cornejo-Bueno, L., Nieto-Borge, J. C., García-Díaz, P., Rodríguez, G., \& Salcedo-Sanz, S. (2016). Significant wave height and energy flux prediction for marine energy applications: A grouping genetic algorithm – Extreme Learning Machine approach. Renewable Energy, 97, 380–389. 
\item De Stefano, C., Fontanella, F., \& Marrocco, C. (2008). A GA-based feature selection algorithm for remote sensing images. Workshops on Applications of Evolutionary Computation, 285–294. Springer.
\item Fukushima, A., Sugimoto, M., Hiwa, S., \& Hiroyasu, T. (2019). Elastic net-based prediction of IFN-β treatment response of patients with multiple sclerosis using time series microarray gene expression profiles. Scientific Reports, 9(1), 1822.
\item Guyon, I., \& Elisseeff, A. (2003). An Introduction to Variable and Feature Selection. Journal of Machine Learning Research (JMLR).
\item Hall, M. A. (1999). Feature selection for discrete and numeric class machine learning.
\item Hong, H., Tsangaratos, P., Ilia, I., Liu, J., Zhu, A.-X., \& Xu, C. (2018). Applying genetic algorithms to set the optimal combination of forest fire related variables and model forest fire susceptibility based on data mining models. The case of Dayu County, China. Science of The Total Environment, 630, 1044–1056. 
\item Hong, J.-H., \& Cho, S.-B. (2006). Efficient huge-scale feature selection with speciated genetic algorithm. Pattern Recognition Letters, 27(2), 143–150.
\item Hu, Z., Bao, Y., Xiong, T., \& Chiong, R. (2015). Hybrid filter-wrapper feature selection for short-term load forecasting. Engineering Applications of Artificial Intelligence.

\item Huang, C.-L., \& Wang, C.-J. (2006). A GA-based feature selection and parameters optimizationfor support vector machines. Expert Systems with Applications, 31(2), 231–240.
\item Jeong, Y.-S., Shin, K. S., \& Jeong, M. K. (2015). An evolutionary algorithm with the partial sequential forward floating search mutation for large-scale feature selection problems. Journal of The Operational Research Society, 66(4), 529–538.
\item Kabir, M. M., Islam, M. M., \& Murase, K. (2010). A new wrapper feature selection approach using neural network. Neurocomputing, 73(16–18), 3273–3283.
\item Kusmec, A., Srinivasan, S., Nettleton, D., \& Schnable, P. S. (2017). Distinct genetic architectures for phenotype means and plasticities in Zea mays. Nature Plants, 3(9), 715–723. 
\item Leardi, R. (2000). Application of genetic algorithm–PLS for feature selection in spectral data sets. Journal of Chemometrics, 14(5‐6), 643–655. 
\item Lee, W. Y., Garrison, S. M., Kawasaki, M., Venturini, E. L., Ahn, B. T., Boyers, R., … Vazquez, J. (2002).  Low‐temperature formation of epitaxial Tl 2 Ca 2 Ba 2 Cu 3 O 10 thin films in reduced O 2 pressure . Applied Physics Letters, 60(6), 772–774.
\item Lin, K.-C., Huang, Y.-H., Hung, J. C., \& Lin, Y.-T. (2015). Feature selection and parameter optimization of support vector machines based on modified cat swarm optimization. International Journal of Distributed Sensor Networks, 11(7), 365869.
\item Liu, S., Tai, H., Ding, Q., Li, D., Xu, L., \& Wei, Y. (2013). A hybrid approach of support vector regression with genetic algorithm optimization for aquaculture water quality prediction. Mathematical and Computer Modelling, 58(3), 458–465.
\item Oztekin, A., Al-Ebbini, L., Sevkli, Z., \& Delen, D. (2018). A decision analytic approach to predicting quality of life for lung transplant recipients: A hybrid genetic algorithms-based methodology. European Journal of Operational Research, 266(2), 639–651. 
\item Park, I. W., \& Mazer, S. J. (2018). Overlooked climate parameters best predict flowering onset: Assessing phenological models using the elastic net. Global Change Biology, 24(12), 5972–5984. 
\item Peng, H., Long, F., \& Ding, C. (2005). Feature selection based on mutual information: criteria of max-dependency, max-relevance, and min-redundancy. IEEE Transactions on Pattern Analysis \& Machine Intelligence, (8), 1226–1238.
\item Sotomayor, G., Hampel, H., \& Vázquez, R. F. (2018). Water quality assessment with emphasis in parameter optimisation using pattern recognition methods and genetic algorithm. Water Research. 
\item Wang, J., Zareef, M., He, P., Sun, H., Chen, Q., Li, H., … Xu, D. (2019). Evaluation of matcha tea quality index using portable NIR spectroscopy coupled with chemometric algorithms. Journal of the Science of Food and Agriculture, 99(11), 5019–5027.
\item Wei, C., Chen, J., Song, Z., \& Chen, C. I. (2019). Adaptive virtual sensors using SNPER for the localized construction and elastic net regularization in nonlinear processes. Control Engineering Practice.
\item Welikala, R. A., Fraz, M. M., Dehmeshki, J., Hoppe, A., Tah, V., Mann, S., … Barman, S. A. (2015). Genetic algorithm based feature selection combined with dual classification for the automated detection of proliferative diabetic retinopathy. Computerized Medical Imaging and Graphics, 43, 64–77.
\item Xue, B., Zhang, M., Browne, W. N., \& Yao, X. (2015). A survey on evolutionary computation approaches to feature selection. IEEE Transactions on Evolutionary Computation, 20(4), 606–626.
\item Yu, L., \& Liu, H. (2003). Feature selection for high-dimensional data: A fast correlation-based filter solution. Proceedings of the 20th International Conference on Machine Learning (ICML-03), 856–863.
\item Zou, H., \& Hastie, T. (2005). Regularization and variable selection via the elastic net. Journal of the Royal Statistical Society: Series B (Statistical Methodology), 67(2), 301–320.

\end{itemize}

\end{document}